\documentclass[letterpaper, 10 pt, conference]{ieeeconf}

\IEEEoverridecommandlockouts                              

\usepackage{graphics} 
\usepackage{epsfig} 
\usepackage{mathptmx} 
\usepackage{times} 
\usepackage{amsmath} 
\usepackage{amssymb}  
\usepackage{multirow}
\usepackage{booktabs}
\usepackage{cite}
\usepackage{url}
\usepackage{hyperref}

\title{\LARGE \bf
FA-BARF: Frequency Adapted Bundle-Adjusting \\ Neural Radiance Fields
}

\author{Rui Qian$^{1}$, Chenyangguang Zhang$^{2}$, Yan Di$^{3}$, \\ Guangyao Zhai$^{3}$, Ruida Zhang$^{2}$, Jiayu Guo$^{1}$, Benjamin Busam$^{3}$, Jian Pu$^{1}$
\thanks{Manuscript received xx xx, 2024. (Corresponding author: Jian Pu).}
\thanks{$^{1}$Rui Qian, Jiayu Guo and Jian Pu are with the Institute of Science and Technology for Brain-Inspired Intelligence, Fudan University, Shanghai 200433, China. E-mails: {\tt\small eleanor\_chien@foxmail.com, jianpu@fudan.edu.cn}.}%
\thanks{$^{2}$Chenyangguang Zhang and Ruida Zhang are with Department of Automation, Tsinghua University, China.}%
\thanks{$^{3}$Yan Di, Guangyao Zhai and Benjamin Busam are with Chair for Computer Aided Medical Procedures and Augmented Reality, Technical University of Munich, Germany. E-mail: {\tt\small b.busam@tum.de}.}%
}

\begin{document}

\maketitle

\thispagestyle{empty}
\pagestyle{empty}



\begin{abstract}

Neural Radiance Fields (NeRF) have exhibited highly effective performance for photorealistic novel view synthesis recently. However, the key limitation it meets is the reliance on a hand-crafted frequency annealing strategy to recover 3D scenes with imperfect camera poses. The strategy exploits a temporal low-pass filter to guarantee convergence while decelerating the joint optimization of implicit scene reconstruction and camera registration. In this work, we introduce the Frequency Adapted Bundle Adjusting Radiance Field (FA-BARF), substituting the temporal low-pass filter for a frequency-adapted spatial low-pass filter to address the decelerating problem. We establish a theoretical framework to interpret the relationship between position encoding of NeRF and camera registration and show that our frequency-adapted filter can mitigate frequency fluctuation caused by the temporal filter. Furthermore, we show that applying a spatial low-pass filter in NeRF can optimize camera poses productively through radial uncertainty overlaps among various views. Extensive experiments show that FA-BARF can accelerate the joint optimization process under little perturbations in object-centric scenes and recover real-world scenes with unknown camera poses. This implies wider possibilities for NeRF applied in dense 3D mapping and reconstruction under real-time requirements. The code will be released upon paper acceptance.
\end{abstract}

\section{Introduction}\label{intro}


In the last few decades, Structure from Motion (SfM) \cite{schonberger2016structure} and visual Simultaneous Localization and Mapping (visual SLAM) \cite{campos2021orb,zhai2020poseconvgru} techniques have gained significant interest from both the computer vision and robotic communities, including a wide range of applications, such as robot navigation \cite{zhang2021efficient} and augmented reality \cite{liu2016robust}. 
As a crucial part of refining a visual reconstruction to produce jointly optimal 3D structure and viewing parameter estimates in SfM and SLAM, classical bundle adjustment is a large sparse geometric parameter estimation problem, the parameters being the combined 3D feature coordinates and camera poses. While NeRF \cite{mildenhall2021nerf} provides a space-efficient implicit neural representation of dense geometric reasoning, bundle adjustment combined with the implicit 3D structure integrates abundant geometric information with a compact memory footprint for downstream vision tasks, which used to be limited by the sparse nature of output 3D point clouds in the classical context. 

Given a collection of images captured by camera sensors, implicit bundle adjustment targets to recover the 3D scene as a neural network mapping 3D features to complex signals (e.g. density or color), which can synthesize images from arbitrary views through volumetric rendering \cite{levoy1990efficient}, and register the corresponding camera poses to locate the ego-motion of sensors. Considering camera poses as independent variables in SE(3), the BARF series methods \cite{lin2021barf,truong2023sparf,chen2023local} render the implicit model of the 3D scene to the observed views through initialized poses, construct photometric error between rendered and ground truth pixels as the loss function, and optimize poses and learnable scene representation jointly.

\begin{figure}[t]
    \centering  \includegraphics[width=1\linewidth,page=1]{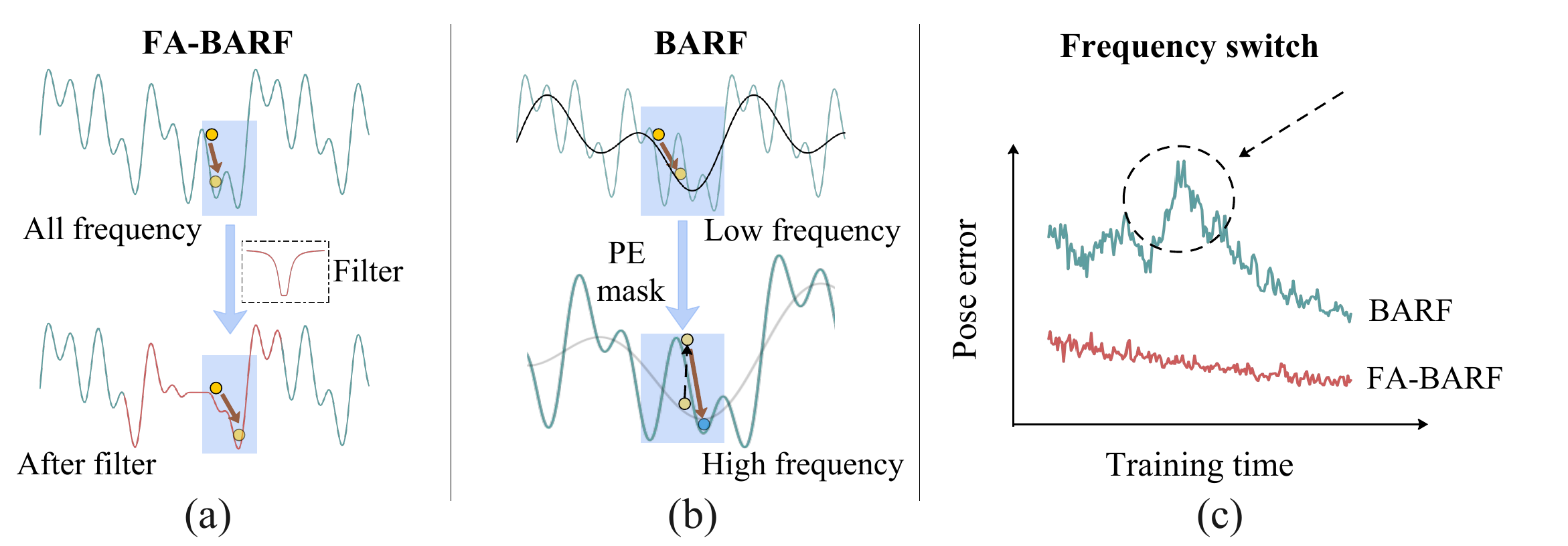}
    \caption{
        Comparision of pose optimization process between FA-BARF and BARF. (a) FA-BARF utilizes a frequency adapted spatial low-pass filter to adjust the ability of optimizing poses among different frequencies. (b) BARF adopts a temporal low-pass filter to guide pose optimization from low frequency to high frequency. (c) The temporal low-pass filter causes frequency fluctuation, impeding the process of pose optimization with frequency switch during the training process.
    }
    \label{fig:teaser}
\end{figure}

Despite BARF's notable ability of reconstruction and registration, the adopted hand-crafted frequency annealing strategy \cite{park2021nerfies} sacrifices the efficiency of the implicit model’s learning process to guarantee the convergence of the algorithm. As illustrated in Fig. \ref{fig:teaser}, BARF applies a smooth mask on the different frequency bands (from low to high) of the implicit model over the course of optimization, acting like a temporal low-pass filter. The temporal filter guides poses from a coarser direction associated with lower frequencies to a finer direction associated with higher frequencies while introducing the frequency fluctuation. The frequency fluctuation means the optimization process of poses is impeded when the learned frequencies are disturbed by higher frequencies joined later. 

To address the decelerated training process and negative optimization impact caused by the temporal low-pass filter, we propose Frequency Adapted Bundle-Adjusting NeRF (FA-BARF), an innovative implicit bundle adjustment transforming the temporal low-pass filter to a frequency-adapted spatial low-pass filter. In this paper, we aim to preclude the negative impact of frequency fluctuation while recovering NeRF from imperfect camera poses. Starting with a theoretical framework, we analyze the influence on pose optimization caused by different frequency components in NeRF representation through position encoding \cite{mildenhall2021nerf}. Furthermore, we show that our frequency-adapted strategy can mitigate frequency fluctuation through substituting the temporal low-pass filter for a frequency-adapted spatial low-pass filter. The proposed spatial low-pass filter also enhances the sensitivity of rendered views related to varying poses and optimizes corresponding poses effectively by introducing radial uncertainty overlap among different views. To this end, we validate that FA-BARF can accelerate pose convergence and NeRF training process under little perturbations in object-centric scenes, and also recover real-world scenes in the form of implicit representation with unknown camera poses.


In summary, we present the following contributions:

\begin{itemize}
    \item We provide a theoretical framework to analyze the relationship between the position encoding and pose optimization, setting a foundation for interpreting the impact of different frequencies in the joint optimization of reconstruction and registration.
    \item We present that the proposed frequency-adapted low-pass filter can guarantee the optimal results of reconstruction and registration by eliminating the frequency fluctuation phenomenon caused by the classical temporal filter and exploiting the radial uncertainty overlap of different views.
    \item Our proposed FA-BARF can curtail more than 50\% of training time, and improve registration accuracy and view synthesis quality, compared to the original BARF in object-centric scenes. In real-world scenes, FA-BARF can also outperform with fewer registration errors and higher perceptual similarity in view synthesis.
\end{itemize}
\section{Related Work}\label{relatedwork}

\textbf{Implicit Bundle-Adjusting Algorithms.} Given a set of input image tracks, bundle adjustment is performed to refine a visual reconstruction to produce jointly optimal structure and viewing parameter estimates \cite{triggs2000bundle} in modern SfM \cite{agarwal2011building} and visual SLAM  systems \cite{campos2021orb,yang2021asynchronous}, which aim to recover the 3D feature from initial noisy or unknown camera poses. As the dawn of the Neural Radiance Field (NeRF) \cite{wang2021nerf} is breaking, the 3D implicit map has been integrated into the framework of bundle adjustment, as an extension of classical direct methods \cite{yin2018geonet}, exploiting photometric consistency to define the loss function. According to different optimization patterns related to camera poses, these implicit bundle adjustment algorithms tilt into two main orientations, (a) global registration \cite{levy2023melon,lin2021barf,wang2021nerf} that optimizes absolute poses consistently and (b) local-to-global registration \cite{cheng2023lu,chen2023local} that optimizes absolute poses and relative poses progressively.

Note that global registration methods are fundamental strategies adopted by local-to-global registration methods in the local optimization phase, our work targets to enhance the accuracy and convergence rate of poses in the SE(3) manifold, providing a better baseline with less time-consuming in different implicit bundle adjustment algorithms.

\textbf{Explicit and Implicit Pose Optimization.} According to different parameterization of camera poses, the pose optimization algorithms can be classified into (a) learning-based methods which train a pose encoder to regress poses from 2D images or 3D geometric features and (b) explicit pose methods which optimizes 6DOF poses directly. Learning-based approaches include GAN-based pose estimation \cite{meng2021gnerf,chan2021pi,nguyen2019hologan,niemeyer2021giraffe}, diffusion-based pose estimation \cite{wang2023posediffusion,zhang2024raydiffusion}, and iterative pose estimation \cite{sinha2023sparsepose, jiang2024forge}, introducing over-parameterized distributed representations to obtain the optimal estimator. To achieve real-time optimization of poses, explicit pose methods related to NeRF focuses mostly on adapting inverse rendering \cite{yen2021inerf} to challenging scenarios like sparse input views \cite{truong2023sparf}, dramatic movement \cite{bian2023nope}, varying background and illumination \cite{boss2022samurai}, and unbounded scenes \cite{meuleman2023progressively}.

However, the inverse rendering methods mostly rely on the coarse-to-fine positional encoding annealing strategy of BARF, sacrificing the learning time of high frequency to gain the convergence of poses. Our method offers a flexible trade-off between pose and NeRF optimization, expanding the possibility of implicit bundle adjustment algorithms in real-time and challenging applications \cite{kong2023vmap}.

\section{Method}\label{method}


We unfold this paper by constructing a theoretical framework to analyze the influence on pose optimization posed by different frequencies of the scene representation. Through numerical methods, we firstly demonstrate that the positional
encoding annealing strategy obtain the convergence of algorithms by a temporal low-pass filter. Then we propose a frequency-adapted spatial low-pass filter to replace the temporal filter and rule out the frequency fluctuation by removing the temporal release process of frequencies.

\subsection{Bundle Adjusting NeRF}


The optimization process of implicit bundle adjustment can boil down to three main phases, camera intrinsic and extrinsic transformation, implicit scene representation as a neural net, and composite volumetric rendering. To analyze the relationship between different frequencies and pose optimization, we focus on the positional encoding mapping stage of the second phase in the following parts.

To obtain the RGB value of a pixel with image coordinate $\mathbf{u}\in\mathbb{R}^2$ through 3D signals distributed in space through NeRF, a set of points need to be sampled along the ray firstly, which starts from the origin of camera center and passes through the corresponding pixel, with a set of depth values $z_1,\cdots, z_N$ in camera coordinate. Through a 6-DoF camera pose $\mathbf{p}\in \text{SE(3)}$ as the extrinsic parameter and a rigid transformation function $W: \mathbb{R}^3\times\text{SE(3)}\rightarrow\mathbb{R}^3$ as the intrinsic and extrinsic mapping, the sampled 3D point $\mathbf{x}$ in camera view space can be mapped to 3D world coordinates so as to obtain corresponding signals (density and RGB) of each sampled point through the evaluation of the network $f$. In the final phase, the volumetric rendering technique aggregates mapped signals distributed along the ray to approximate the RGB value $\hat{I}$ of a specific pixel. Normally, the whole process can be described as the following equation
\begin{equation}
\hat{I}(\mathbf{u} ; \mathbf{p})=g\left(f\left(W\left(z_1 \overline{\mathbf{u}} ; \mathbf{p}\right) ; \boldsymbol{\Theta}\right), \ldots, f\left(W\left(z_N \overline{\mathbf{u}} ; \mathbf{p}\right) ; \boldsymbol{\Theta}\right)\right),
\label{eq:volumetric_rendering}
\end{equation}
where $\overline{\mathbf{u}}\in\mathbb{R}^3$ represents the homogeneous coordinates of $\mathbf{u}$, $g: \mathbb{R}^{4N}\rightarrow\mathbb{R}^3$ represents the ray rendering function, and $\boldsymbol{\Theta}$ represents the parameters of network $f$. 

The ultimate target of bundle adjusting NeRF is to optimize the parameters of the network $\boldsymbol{\Theta}$ and camera poses $\mathbf{p}$ jointly under the supervision of RGB values from $M$ different views. Therefore, our optimization framework has the following form,
\begin{equation}
\min _{\mathbf{p}_1, \ldots, \mathbf{p}_M, \boldsymbol{\Theta}} \sum_{i=1}^M \sum_{\mathbf{u}}\left\|\hat{I}\left(\mathbf{u} ; \mathbf{p}_i, \boldsymbol{\Theta}\right)-I_i(\mathbf{u})\right\|_2^2,
\label{eq:loss}
\end{equation}
where $I_i$ denotes the real RGB value of the $i$-th captured camera view corresponding to the pixel $\mathbf{u}$.

Furthermore, according to the backpropagation process accomplished in practice, gradient-based optimization points out $\mathbf{p}$ can be updated through $\mathbf{p} \leftarrow \mathbf{p}+\Delta \mathbf{p}$, and the updates of poses has the form 
\begin{equation}
\Delta \mathbf{p}=-\mathbf{A}(\mathbf{u} ; \mathbf{p}, \boldsymbol{\Theta}) \sum_{\mathbf{u}} \mathbf{J}(\mathbf{u} ; \mathbf{p}, \boldsymbol{\Theta})^{\top}\Delta I,
\end{equation}
where $\mathbf{A}$ is a generic matrix which depends on the choice of the optimization algorithm, and the Jacobian matrix $\mathbf{J}$ demonstrates the convergence tendency of camera poses related to the photometric loss $\Delta I$ between approximated RGB values $\hat{I}$ and observed RGB values $I$ as defined in Eq. (\ref{eq:loss}). The Jacobian matrix $\mathbf{J}$ can be expanded as
\begin{equation}
\mathbf{J}(\mathbf{u} ; \mathbf{p}, \boldsymbol{\Theta})=\sum_{i=1}^N \frac{\partial g\left(\mathbf{y}_1, \ldots, \mathbf{y}_N\right)}{\partial \mathbf{y}_i} \frac{\partial \mathbf{y}_i(\mathbf{p},\boldsymbol{\Theta})}{\partial \mathbf{x}_i(\mathbf{p})} \frac{\partial W\left(z_i \overline{\mathbf{u}} ; \mathbf{p}\right)}{\partial \mathbf{p}},
\label{eq:jacobian}
\end{equation}
where $\mathbf{y}_i$ is a four-dimensional signal vector including color and density value corresponding to a sampled point. The Jacobian matrix is composed of three parts, volumetric rendering, network mapping,  extrinsic and intrinsic transformation in an inverted order, corresponding to the three main phases in the rendering process. 
Based on this theoretical framework, we are able to analyze the relationship between pose optimization and the frequency of implicit scene representation in the next part.

\subsection{Pose Optimization Analysis}

Multi Layer Perceptrons (MLP) are a crucial part of NeRF, which map low dimensional position points $\mathbf{x}\in \mathbb{R}^{3}$ to output values of signals with high frequency. Considering the conventional MLP with ReLU exhibiting a deficient pattern of spectral bias \cite{rahaman2019spectral}, various position encoding method has been introduced as a pre-embedding strategy to mitigate this biased learning problem by projecting the inputs into a higher dimensional space through a set of sinusoids.

Position encoding is commonly described as $\gamma: \mathbb{R}^3 \rightarrow \mathbb{R}^{3+6 L}$ with $L$ frequency denoted as
\begin{equation}
\gamma(\mathbf{x})=\left[\mathbf{x}^{\mathrm{T}}, \gamma_0(\mathbf{x}), \gamma_1(\mathbf{x}), \ldots, \gamma_{L-1}(\mathbf{x})\right] \in \mathbb{R}^{3+6 L},
\end{equation}
where the $k$-th frequency basis $\gamma_k$ is
\begin{equation}
\gamma_k(\mathbf{x})=\left[\sin \left(2^k \mathbf{x}^{\mathrm{T}}\right), \cos \left(2^k \mathbf{x}^{\mathrm{T}}\right)\right] \in \mathbb{R}^6,
\end{equation}
with the sinusoidal function set operating coordinate-wise. It is worthy to notice that the input of MLP in NeRF has been lifted to $\gamma_k(\mathbf{x})$, as the substitution of original 3D points with abundant frequency expression. 

In this case, the mathematical expression of network $f$ can be rewritten as $ f^{'}(\gamma(\mathbf{x}))$, where $ f^{'}$ denotes the main network structure of $f$. The Jacobian matrix of poses related to the neural net has the form of $\partial (f^{'}(\gamma))/\partial \mathbf{p}$, which is equal to $(\partial f^{'} /\partial \gamma)\cdot(\partial \gamma/\partial \mathbf{p})$ according to the chain rules. To analyze the relationship between $\gamma$ and camera poses $\mathbf{p}$, we derive the Jacobian matrix of camera poses related to different frequency components as
\begin{equation}
\begin{aligned}
    \frac{\partial \gamma_k(\mathbf{x})}{\partial \mathbf{d}_w} 
    &= \left[\begin{array}{c}
        2^k\cdot\cos \left(2^k\mathbf{x}\right)\odot \mathbf{I}_3\\
        -2^k\cdot \sin \left(2^k\mathbf{x}\right)\odot \mathbf{I}_3\end{array}\right] \cdot x_t,
        \\ 
    \frac{\partial \gamma_k(\mathbf{x})}{\partial \mathbf{t}_{c2w}} &=\left[\begin{array}{c}
    2^k\cdot\cos \left(2^k\mathbf{x}\right)\odot \mathbf{I}_3  \\
    -2^k\cdot \sin \left(2^k\mathbf{x}\right)\odot \mathbf{I}_3
        \end{array}\right],
    \end{aligned}
\label{eq:jacobianoriginalpe}
\end{equation}
where $x_t$ denotes the distance from camera center to a sampled 3D point, $\mathbf{d}_w$ denotes the direction of the a sampled ray in world coordinate, encoding the rotation of $\mathbf{p}$, $\mathbf{t}_{c2w}$ denotes the translation of $\mathbf{p}$ in the world coordinate, $\mathbf{I}_3$ is the identity matrix with dimensions three, the symbol $\odot$ represents element-wise multiplication, and $\odot\mathbf{I}_3$ denotes expanding a three-dimensional vector to a three-dimensional diagonal matrix. 

As demonstrated in \cite{lin2021barf}, the positional encoding mapping leads to sub-optimal solutions of bundle adjustment. Thus BARF \cite{lin2021barf} adopted a coarse-to-fine
positional encoding annealing strategy \cite{park2021nerfies} to address this problem, adding frequency components from low to high gradually during the training process. From Eq. (\ref{eq:jacobianoriginalpe}), we can observe that the core idea of BARF is trusting low frequency first and then fixing pose optimization direction in details according to high frequency information progressively, which acts like a temporal low-pass filter. Although the progressive position encoding mask can guarantee the convergence of bundle adjustment, the temporal low-pass filter introduces frequency fluctuation that causes mutual interference among dynamic frequencies during the training process, as shown in Fig. \ref{fig:teaser}.


\subsection{Adaptive Pose Optimization}

\begin{figure}[t!]
    \centering  
    \includegraphics[width=1\linewidth,page=1]{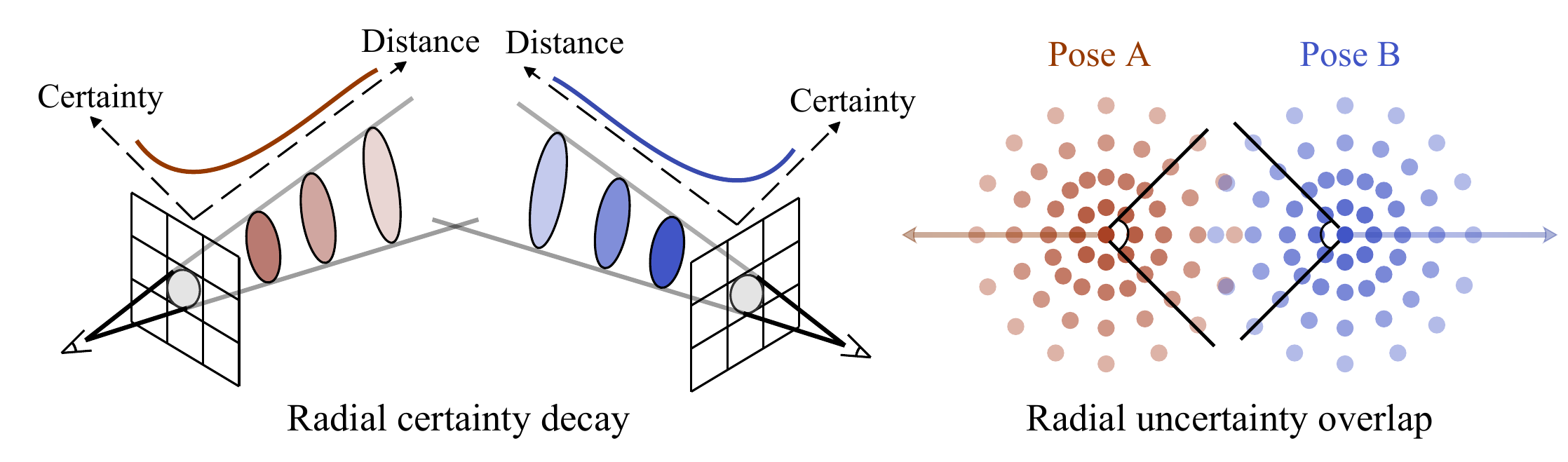}
    \caption{
        Visual interpretation of radial certainty overlaps related to camera poses. As defined in \cite{barron2021mip}, the covariance of sampled points with surrounding Gaussian region decreases when the distance between camera center and sampled point decreases with higher certainty to adjust the orientation of pose optimization. The shade of colour represents the degree of certainty. The deeper colour denotes higher certainty of the sampled point.
    }
    \label{fig:radialcertainty}
    \vspace{-8pt}
\end{figure}

To mitigate the frequency fluctuation caused by the temporal low-pass filter, we adopt a frequency-adapted spatial strategy to adjust the impact that the various frequency signals exert on pose optimization with Integrated Position Encoding (IPE) \cite{barron2021mip}. IPE introduced the cone sampling strategy to encode a 3D point and its surrounding Gaussian region, transforming the original position encoding into the integrated position encoding. It contains the mean and covariance information related to the sampled 3D cone frustum, described as
\begin{equation}
\bar{\gamma}(\boldsymbol{\mu}, \boldsymbol{\Sigma})=\left[\boldsymbol{\mu}^{\mathrm{T}}, \bar{\gamma}_0(\boldsymbol{\mu}, \boldsymbol{\Sigma}), \bar{\gamma}_1(\boldsymbol{\mu}, \boldsymbol{\Sigma}), \ldots, \bar{\gamma}_{L-1}(\boldsymbol{\mu}, \boldsymbol{\Sigma})\right] \in \mathbb{R}^{3+6 L},
\end{equation}
where $\boldsymbol{\mu}$ and $\boldsymbol{\Sigma}$ represent the mean and covariance of conical frustum as the multivariate Gaussian form respectively, with the explicit expression of 
\begin{equation}
\boldsymbol{\mu}=\mathbf{o}+\mu_t \mathbf{d}_w, \quad \boldsymbol{\Sigma}=\sigma_t^2\left(\mathbf{d}_w \mathbf{d}_w^{\mathrm{T}}\right)+\sigma_r^2\left(\mathbf{I}-\frac{\mathbf{d}_w \mathbf{d}_w^{\mathrm{T}}}{\|\mathbf{d}_w\|_2^2}\right),
\end{equation}
where $\mathbf{o}$ denotes camera's center, $\mathbf{d}_w$ denotes the direction of the casting ray in world coordinates, $\mu_t$ denotes the mean distance between camera center and sampled point along the ray, $\sigma_t^2$ and $\sigma_r^2$ denote the variance information along and perpendicular to the ray respectively.

The expression of integrated position encoding feature is computing the expectation over the multivariate Gaussian lifted by the set of sinusoids. The expectation of the $k$-th frequency basis has a closed-form expression as
\begin{equation}
\begin{aligned}
\bar{\gamma}_{k}(\boldsymbol{\mu}, \boldsymbol{\Sigma}) & =\mathrm{E}_{\mathbf{x} \sim \mathcal{N}\left(\boldsymbol{\mu}, \boldsymbol{\Sigma}\right)}[\bar{\gamma}_k(\mathbf{x})] \\
& =\left[\begin{array}{l}
\sin \left(2^k\boldsymbol{\mu}\right) \cdot \exp \left(-\frac{1}{2} \cdot 4^k \cdot \operatorname{diag}\left(\boldsymbol{\Sigma}\right)\right) \\
\cos \left(2^k\boldsymbol{\mu}\right) \cdot \exp \left(-\frac{1}{2} \cdot 4^k \cdot \operatorname{diag}\left(\boldsymbol{\Sigma}\right)\right)
\end{array}\right]^{\mathrm{T}}  \in \mathbb{R}^6,
\end{aligned}
\label{eq:ipe}
\end{equation}
which constitutes the input components of the MLP. To analyze the relationship between integrated position encoding $\bar{\gamma}_k$ and camera poses $\mathbf{p}$, we derive the Jacobian of rotation part $\mathbf{d}_w$ and translation part $\mathbf{t}_{c2w}$ related to $\bar{\gamma}_k$ similarly as
\begin{equation}
\resizebox{1.0\hsize}{!}{
$\begin{aligned}
\frac{\partial \bar{\gamma}_k(\boldsymbol{\mu}, \boldsymbol{\Sigma})}{\partial \mathbf{d}_w} 
&\sim \left[\begin{array}{c}
    2^k\cdot\cos \left(2^k\boldsymbol{\mu}\right) \cdot \exp \left(-\frac{1}{2} \cdot 4^k \cdot\operatorname{diag}\left(\boldsymbol{\Sigma}\right)\right)\odot \mathbf{I}_3\\
    -2^k\cdot \sin \left(2^k\boldsymbol{\mu}\right) \cdot \exp \left(-\frac{1}{2} \cdot 4^k \cdot\operatorname{diag}\left(\boldsymbol{\Sigma}\right)\right)\odot \mathbf{I}_3
    \end{array}\right] \cdot \mu_t,
    \\ 
    \frac{\partial \bar{\gamma}_k(\boldsymbol{\mu}, \boldsymbol{\Sigma})}{\partial \mathbf{t}_{c2w}} &=\left[\begin{array}{c}
    2^k\cdot\cos \left(2^k\boldsymbol{\mu}\right) \cdot \exp \left(-\frac{1}{2} \cdot 4^k \cdot\operatorname{diag}\left(\boldsymbol{\Sigma}\right)\right)\odot \mathbf{I}_3\\
    -2^k\cdot \sin \left(2^k\boldsymbol{\mu}\right) \cdot \exp \left(-\frac{1}{2} \cdot 4^k \cdot\operatorname{diag}\left(\boldsymbol{\Sigma}\right)\right)\odot \mathbf{I}_3
    \end{array}\right],
\end{aligned}$
}
\label{eq:jacobianipe}
\end{equation}
where the $\sim$ represents an approximated operation. This equation is explained as an extended derivation in the Appendix. 

Compared to Eq. (\ref{eq:jacobianoriginalpe}), Eq. (\ref{eq:jacobianipe}) multiplies the Jacobian coefficient of $k$-th frequency basis with exponential constants including the $k$-th power of four and the covariance matrix information of sampled cones. On the one hand, higher frequency components embrace exponential parts which are closer to zero, decreasing the impact of high frequency components on pose optimization. With the adjustment targeting to different frequencies, the joint optimization of reconstruction and registration can transform the temporal low-pass filter to a constant frequency-adapted low-pass filter on positional encoding, avoiding the frequency fluctuation phenomenon caused by the dynamic positional encoding mask. 

On the other hand, the spatial low-pass filter leads to higher sensitivity of pose optimization through integrating the covariance information of sampled points with surrounding Gaussian region. As shown in Fig. \ref{fig:radialcertainty}, each pose embraces a radial uncertainty field defined by the distance between sampled points and the camera center. The error of registration will be effectively reflected and optimized by the loss between observed and rendered views, especially when the sampled points fall into radial uncertainty overlaps among the various views. Therefore, the proposed strategy guarantees the convergence and effectiveness of NeRF bundle adjustment with a) the constant frequency-adapted filter to balance the impact of different frequencies exerted on pose optimization, and b) the radial uncertainty field to update poses through the covariance information of sampled points with surrounding Gaussian region under various views.

\section{Experimental Results}\label{exp}

We validate the effectiveness of our proposed FA-BARF with an object-centric dataset and a real-world dataset, showing how the adaptive pose optimization strategy can be generalized to implicit bundle adjustment algorithms.

\subsection{Synthetic Objects}\label{sec:blender-expset}

To demonstrate the impact of our frequency-adapted positional encoding strategy in implicit reconstruction from imperfect camera poses, we experiment with the eight synthetic object-centric scenes provided by \cite{mildenhall2021nerf}, which consists of $M = 100$ rendered images with groundtruth camera poses for each scene for training.

\subsubsection{Experimental settings}
The camera poses $\mathbf{p}$ are parameterized with the SE(3) Lie algebra and assume known intrinsics provided by dataset. For each scene, we synthetically perturb the camera poses with additive noise. Following BARF~\cite{lin2021barf}, we chose a standard deviation of $14.9^{\circ}$ in rotation and 0.26 in translational magnitude. We then optimize the scene representation and the camera poses jointly. We evaluate FA-BARF mainly against the original BARF model with or without the coarse-to-fine positional encoding mask. 

\subsubsection{Implementation details}
We follow the architectural settings from  \cite{mildenhall2021nerf} with some modifications.
We train a single MLP with 128 hidden units in each layer and without additional hierarchical sampling for simplicity.
We resize the images to $400 \times 400$ pixels and randomly sample 1024 pixel rays at each optimization step. We choose $N = 128$ sample for numerical integration along each ray, and we use the softplus activation on the volume density output $\sigma$ for improving stability. 
We use the Adam optimizer and train all models for 200K iterations, with a learning rate of $5\times10^{-4}$ exponentially decaying to $1\times10^{-4}$ for the network f and $1\times10^{-3}$ decaying to $1\times10^{-5}$ for the poses $\mathbf{p}$. For BARF, we linearly adjust $\alpha$ from iteration 20K to 100K and activate all frequency bands (up to $L = 10$) subsequently. For FA-BARF, we abandon the position encoding mask to validate our adaptive frequency assumption.

\subsubsection{Evaluation criteria}
We measure the performance in four aspects: pose error  and convergence speed for registration, and view synthesis quality and training speed for the scene representation. Since both the scene and camera poses are variable up to a 3D similarity transformation \cite{lin2021barf}, we evaluate the quality of registration by pre-aligning the optimized poses to the ground truth with Procrustes analysis on the camera locations. For evaluating view synthesis, we run an additional step of test-time photometric optimization on the trained models \cite{lin2019photometric,yen2021inerf} to factor out the pose error that may contaminate the view synthesis quality. We report the average rotation and translation errors for pose and PSNR, SSIM and LPIPS \cite{zhang2018unreasonable} for view synthesis as indices to evaluate the performance of different algorithms. For evaluating the speed of pose convergence, we record the training time when the translation error is lower than $1\times 10^{-2}$ and $5\times 10^{-3}$ in magnitude, and the rotation error is lower than $0.29^{\circ}$ (around $5\times 10^{-3}$ in radian measure). For evaluating the training speed of scene representation, we record PSNR values of rendered views in test datasets at 0, 20, 40, 180, 360 minutes after the beginning of training.

\begin{figure}[t]
    \centering  
    \includegraphics[width=1\linewidth,page=1]{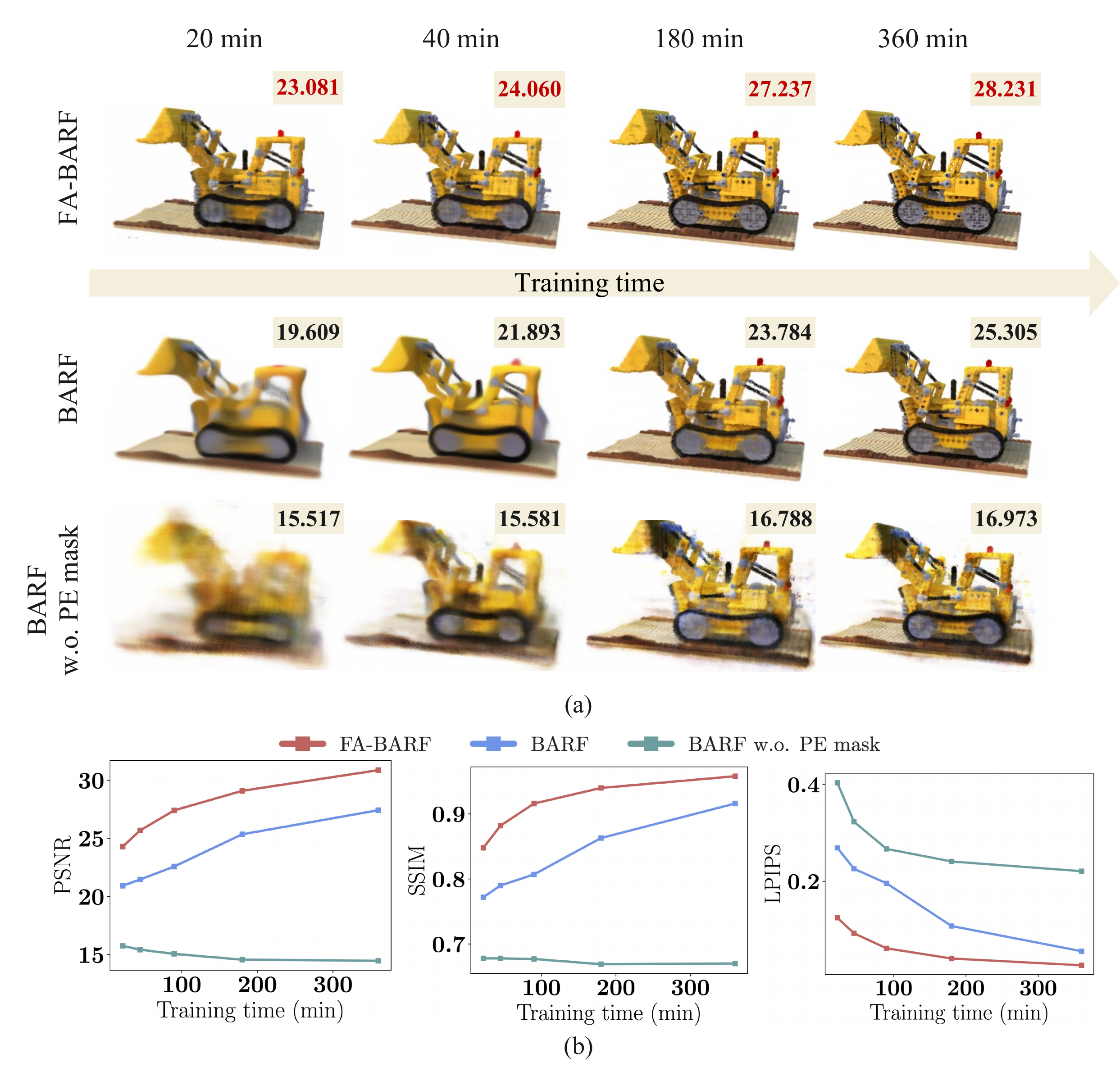}
    
    \caption{
        Visual accelerated reconstruction related to FA-BARF and BARF for the \textit{lego} scene. (a) compares the PSNR index with visual demonstration of view synthesis among BARF without positional encoding mask, original BARF and FA-BARF as training time increases. (b) compares PSNR, SSIM and LPIPS among the three settings with increasing training time. FA-BARF achieves the best performance in reconstruction during the same time compared to original BARF, while BARF gets stuck in sub-optimal results without the positional encoding mask.
    }
    \label{fig:blender_acceleration}
    \vspace{-8pt}
\end{figure}

\begin{figure}[t]
    \centering  \includegraphics[width=0.9\linewidth,page=1]{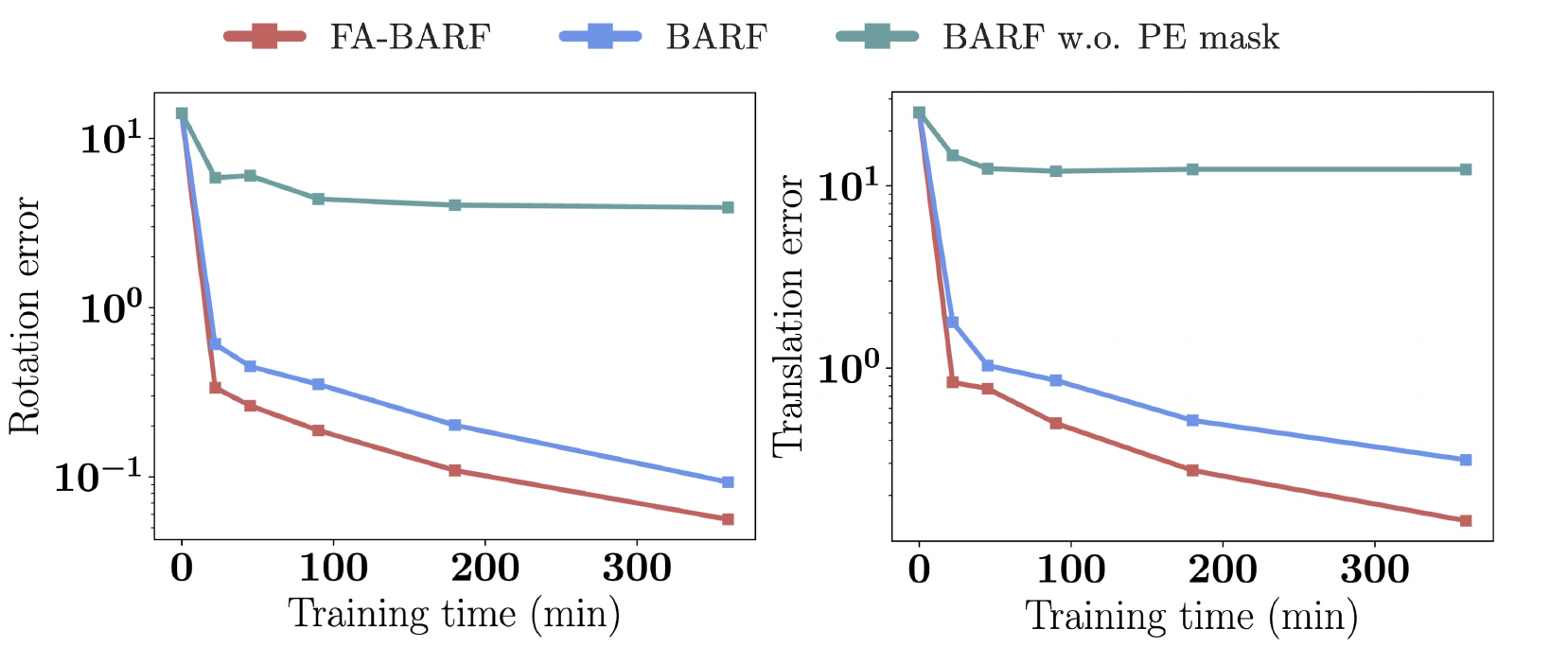}
    \caption{
        Visual accelerated registration related to FA-BARF and BARF for the \textit{lego} scene. FA-BARF assures the convergence of camera poses faster than original BARF, while poses diverge to sub-optimal results in BARF without the positional encoding mask. The rotation errors are in degree and the translation errors are scaled by 100.
    }
    \label{fig:pose_error_acceleration}
    \vspace{-8pt}
\end{figure}

\begin{table}[t]
\caption{Comparision of pose convergence speed related to FA-BARF and BARF. Translation errors are scaled by 100.
    }
    \vspace{-16pt}
\label{tab:GT_model}
\begin{center}
\resizebox{\linewidth}{!}{
\begin{tabular}{l|@{~}c|c|c@{~}}
\toprule
Method          & $\text{Rotation}<0.29^{\circ} \downarrow$         & $\text{Translation}<1.00 \downarrow$              & $\text{Translation}<0.50 \downarrow$\\ \toprule
BARF (with PE mask)           & 90 min                                 & 50 min                                 & 140 min  \\
FA-BARF (without PE mask)        & \bf{40 min}                            & \bf{12 min}                            & \bf{60 min}  \\      
\bottomrule     
\end{tabular}%
}
\end{center}
\vspace{-8pt}
\label{table:blender_acceleration}
\end{table}

\begin{figure}[t!]
    \centering  
    \includegraphics[width=1\linewidth,page=1]{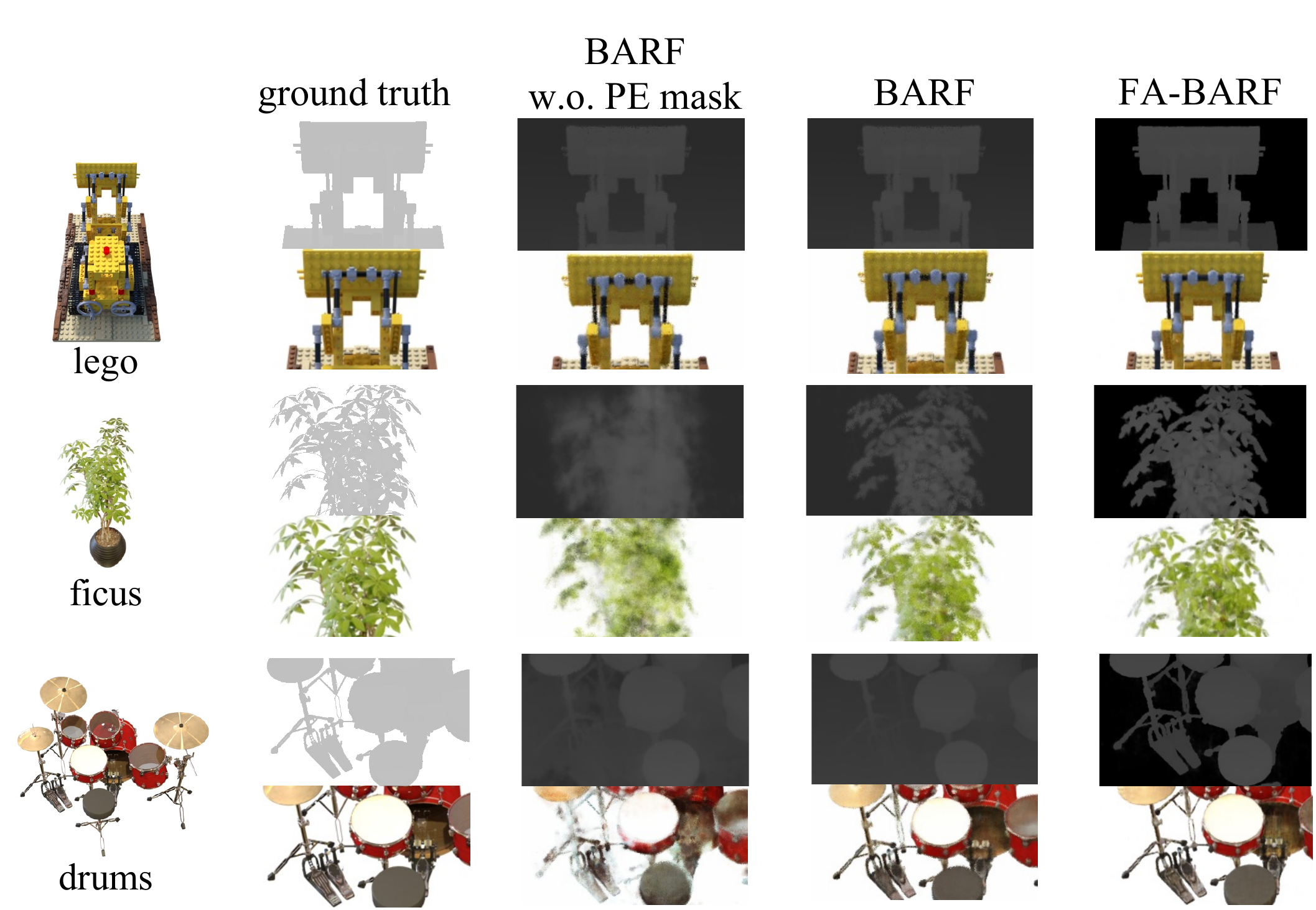}
    \caption{
        Qualitative results of FA-BARF and BARF on synthetic scenes. We visualize the expected depth through ray compositing (top) and the image synthesis (bottom). FA-BARF achieves the best synthesis view quality without PE mask, while original BARF  results in suboptimal registration without PE mask, leading to synthesis artifacts.
    }
    \label{fig:nerf-blender}
    \vspace{-8pt}
\end{figure}

\begin{table*}[t!]
    \caption{
        Quantitative results of FA-BARF and BARF on synthetic scenes. Translation errors are scaled by 100.
    }
    \centering
    \resizebox{\linewidth}{!}{
        \begin{tabular}{l||ccc|ccc||ccc|ccc|ccc}
            \toprule
            \multirow{4}{*}{Scene} & \multicolumn{6}{c||}{Camera pose registration} & \multicolumn{9}{c}{View synthesis quality} \vspace{1.5pt} \\
            & \multicolumn{3}{c|}{Rotation  $\downarrow$} & \multicolumn{3}{c||}{Translation $\downarrow$} & \multicolumn{3}{c|}{PSNR $\uparrow$} & \multicolumn{3}{c|}{SSIM $\uparrow$} & \multicolumn{3}{c}{LPIPS $\downarrow$} \\
            \cmidrule{2-16}
            & \small BARF & \small \multirow{2}{*}{BARF} & \small \multirow{2}{*}{FA-BARF}
            & \small BARF  & \small \multirow{2}{*}{BARF}& \small \multirow{2}{*}{FA-BARF}
            & \small BARF & \small \multirow{2}{*}{BARF} & \small \multirow{2}{*}{FA-BARF}
            & \small BARF & \small \multirow{2}{*}{BARF} & \small \multirow{2}{*}{FA-BARF}
            & \small BARF & \small \multirow{2}{*}{BARF} & \small \multirow{2}{*}{FA-BARF} \vspace{-2.5pt} \\
            & \small  w/o mask& \small & \small 
            & \small  w/o mask & \small & \small 
            & \small  w/o mask & \small & \small 
            & \small  w/o mask & \small & \small 
            & \small  w/o mask & \small & \small  \\
            \midrule
            Chair      & 7.186 & 0.096 & \bf 0.094& 16.638 & \bf 0.428 & 0.581    & 19.02 & 31.16 & \bf 36.83 & 0.804 & 0.954 & \bf 0.990 & 0.223 & 0.044 & \bf 0.010\\
            Drums      & 3.208 & 0.043 & \bf 0.033& 7.322  & 0.225     & \bf 0.196& 20.83 & 23.91 & \bf 26.90 & 0.840 & 0.900 & \bf 0.920     & 0.166 & 0.099 & \bf 0.060\\
            Ficus      & 9.368 & 0.085 & \bf 0.064& 10.135 & 0.474     & \bf 0.358& 19.75 & 26.26 & \bf 29.38 & 0.836 & 0.934 & \bf 0.960     & 0.182 & 0.058 & \bf 0.030\\
            Hotdog   & 3.290 & 0.248 & \bf 0.177& 6.344  & 1.308     & \bf 1.152& 28.15 & 34.54 & \bf 36.21 & 0.923 & 0.970 & \bf 0.980     & 0.083 & 0.032 & \bf 0.020\\
            Lego       & 3.252 & 0.082 & \bf 0.049& 4.841  & 0.291     & \bf 0.203& 24.23 & 28.33 & \bf 29.83 & 0.876 & 0.927 & \bf 0.960 & 0.102& 0.050 & \bf 0.030\\
            Materials  & 6.971 & 0.844 & \bf 0.667& 15.188 & 2.692     & \bf 2.109& 16.51 & \bf 27.84 & 27.46 & 0.747 & 0.936 &\bf 0.940     & 0.294 & 0.058 & \bf 0.030\\
            Mic        & 10.554& 0.071 & \bf 0.043& 22.724 & 0.301     & \bf 0.156& 15.10 & 31.18 & \bf 33.20 & 0.788 & 0.969 & \bf 0.970     & 0.334 & 0.048 & \bf 0.040\\
            Ship       & 5.506 & \bf 0.075 & 0.090& 7.232  & \bf 0.326     & 0.595& 22.12 & 27.50 & \bf 29.08 & 0.755 & \bf 0.849 & 0.810     & 0.255 & \bf 0.132 & 0.140\\
            \midrule
            Average       & 6.167 & 0.193 & \bf 0.152& 11.303 &0.756  &\bf 0.669 & 22.12 & 27.50 & \bf 31.11 & 0.821 &  0.930 & \bf 0.941 & 0.205     & 0.065& \bf 0.045 \\
            \bottomrule 
        \end{tabular}
    }
    \label{table:nerf-blender}
\end{table*}

\begin{table*}[t]
    \caption{
            Quantitative results of FA-BARF and BARF without the coarse-to-fine positional encoding strategy on the LLFF forward-facing scenes from \emph{unknown} camera poses. Translation errors are scaled by 100.
            }
    \centering
        \resizebox{\linewidth}{!}{
            \begin{tabular}{c||cc|cc||cc|cc|cc}
                \toprule
                \multirow{4}{*}{Scene} & \multicolumn{4}{c||}{Camera pose registration} & \multicolumn{6}{c}{View synthesis quality} \vspace{1.5pt} \\
                & \multicolumn{2}{c|}{Rotation (degree) $\downarrow$} & \multicolumn{2}{c||}{Translation $\downarrow$} & \multicolumn{2}{c|}{PSNR $\uparrow$} & \multicolumn{2}{c|}{SSIM $\uparrow$} & \multicolumn{2}{c}{LPIPS $\downarrow$} \\
                \cmidrule{2-11}
                & \small BARF & \small FA-BARF
                & \small BARF & \small FA-BARF
                & \small BARF & \small FA-BARF
                & \small BARF & \small FA-BARF
                & \small BARF & \small FA-BARF
                \\
                & \small w/o mask& \small w/o mask
                & \small w/o mask& \small w/o mask
                & \small w/o mask& \small w/o mask
                & \small w/o mask& \small w/o mask
                & \small w/o mask& \small w/o mask \\
                \midrule
                Fern       & 74.452  & \bf 0.927& 30.167  & \bf 0.432& 9.81 & \bf 23.33& 0.187& \bf 0.730& 0.853& \bf 0.230\\
                Flower     & 2.525  & \bf 2.453& 2.635  & \bf 0.513& 17.08 & \bf 23.45& 0.344 & \bf 0.690& 0.490 &\bf  0.160\\
                Fortress   &  75.094  & \bf 1.125& 33.231  & \bf 0.951&12.15 & \bf 28.05& 0.270 & \bf 0.760 & 0.807 &\bf 0.220 \\
                Horns     & 58.764  & \bf 5.113 & 32.664  & \bf 2.419& 8.89 & \bf 19.79& 0.158 & \bf 0.650 & 0.805 & \bf 0.330 \\
                Leaves    & 88.091  & \bf 2.105& 13.540  & \bf 0.480& 9.64 & \bf 16.98 & 0.067 & \bf 0.480 & 0.782 & \bf 0.310 \\
                Orchids    & 37.104  & \bf 1.407& 20.312  & \bf 0.820& 9.42 & \bf 17.44 & 0.085 & \bf 0.520 & 0.806 & \bf 0.220 \\
                Room       & 173.811  & \bf 0.420& 66.922  & \bf 0.322&  10.78 & \bf 31.80& 0.278 & \bf 0.950 & 0.871 & \bf 0.090 \\
                T-rex      & 166.231  & \bf 0.563& 53.309  & \bf 0.430& 10.48 & \bf 21.55 &0.158 & \bf 0.740 & 0.885 & \bf 0.250 \\
                \midrule
                Average       & 84.509  & \bf 1.764 & 31.598  & \bf 0.796& 11.03 & \bf 22.80 & 0.193 & \bf 0.690 & 0.787 & \bf 0.226 \\
                \bottomrule 
            \end{tabular}
        }
        \label{table:llff-withoutpemask}
\end{table*}

\begin{table*}[t]
\caption{
        Quantitative results of FA-BARF and BARF with the coarse-to-fine positional encoding strategy on the LLFF forward-facing scenes from \emph{unknown} camera poses. Translation errors are scaled by 100.
        }
\centering
    \resizebox{\linewidth}{!}{
        \begin{tabular}{c||cc|cc||cc|cc|cc}
            \toprule
            \multirow{3}{*}{Scene} & \multicolumn{4}{c||}{Camera pose registration} & \multicolumn{6}{c}{View synthesis quality} \vspace{1.5pt} \\
            & \multicolumn{2}{c|}{Rotation (degree) $\downarrow$} & \multicolumn{2}{c||}{Translation $\downarrow$} & \multicolumn{2}{c|}{PSNR $\uparrow$} & \multicolumn{2}{c|}{SSIM $\uparrow$} & \multicolumn{2}{c}{LPIPS $\downarrow$} \\
            \cmidrule{2-11}
            & \small BARF & \small FA-BARF
            & \small BARF& \small FA-BARF
            & \small BARF & \small FA-BARF
            & \small BARF & \small FA-BARF
            & \small BARF & \small FA-BARF\\
            & \small w/ mask& \small w/ mask
            & \small w/ mask& \small w/ mask 
            & \small w/ mask& \small w/ mask
            & \small w/ mask& \small w/ mask 
            & \small w/ mask& \small w/ mask \\
            \midrule
            Fern       & 0.191  & \bf 0.188& \bf 0.192  & 0.198& \bf 23.79 & 23.66& 0.710& \bf 0.720& 0.311& \bf 0.260\\
            Flower     & 0.251  & \bf 0.182& \bf 0.224  & 0.232& \bf 23.37 & 22.93& \bf 0.698 & 0.670& 0.211 &\bf  0.200\\
            Fortress   &  0.479  & \bf 0.429& 0.364  & \bf 0.362&\bf 29.08 & 28.96& 0.823 & \bf 0.830 & 0.132 & \bf 0.120 \\
            Horns      & \bf 0.304  & 0.335 & 0.222  & \bf 0.186& 22.78 & \bf 23.29& 0.727 & \bf 0.750 & 0.298 & \bf 0.230 \\
            Leaves     & 1.272  & \bf 1.029& \bf 0.249  & 0.273& \bf 18.78 & 17.77 & \bf 0.537 & 0.490 & 0.353 & \bf 0.320 \\
            Orchids    & 0.627  & \bf 0.575& 0.404  & \bf 0.385& \bf 19.45 & 19.20 & \bf 0.574 & 0.570 & 0.291 & \bf 0.280 \\
            Room       & 0.320  & \bf 0.319& 0.270  & \bf 0.268&  31.95 & \bf 32.11& 0.940 & \bf 0.950 & 0.099 & \bf 0.070 \\
            T-rex      & 1.138  & \bf 0.523& 0.720  & \bf 0.431& 22.55 & \bf 22.71 &\bf 0.767 & 0.760 & 0.206 & \bf 0.190 \\
            \midrule
            Average       & 0.573  & \bf 0.455 & 0.331  & \bf 0.289& \bf 23.97 & 23.83 & \bf 0.722 & 0.716 & 0.238 & \bf 0.208 \\
            \bottomrule 
        \end{tabular}
    }
    \label{table:llff-withpemask}
\end{table*}

\subsubsection{Results}
We compare the training speed related to scene representation in Fig. \ref{fig:blender_acceleration}. FA-BARF can achieve high view synthesis quality with structure details in 20 minutes, while BARF costs 180 minutes to learn comparable implicit models with enough frequency scope as the coarse-to-fine positional encoding mask unlocks higher frequency bands. As the training time increases, FA-BARF keeps a high training speed until the implicit scene representation converges to a stable NeRF model. Through substituting the temporal filter for our frequency-adapted spatial filter, FA-BARF opens all frequency throughout the training process and optimizes poses effectively while BARF fails without the coarse-to-fine positional encoding mask, as shown in Fig. \ref{fig:pose_error_acceleration}. With the accelerated training and pose optimization process, FA-BARF can curtail more than $50\%$ training time of original BARF while obtaining high accuray of camera poses and quality of view synthesis, as shown in Table \ref{table:blender_acceleration}. The final quantitative results are reported in Table \ref{table:nerf-blender}. The coarse-to-fine position encoding strategy is necessary for BARF to rule out suboptimal results, while FA-BARF can achieve better performance in both pose registration and reconstruction fidelity without the aid of coarse-to-fine position encoding strategy, represented as the qualitative results in Fig. \ref{fig:nerf-blender}.

\subsection{Real-World Scenes}

We investigate the challenging problem of learning neural 3D representations with NeRF on real-world scenes, where the camera poses are unknown. We consider the LLFF dataset \cite{mildenhall2019local}, which consists of eight forward-facing scenes with RGB images sequentially captured by hand-held cameras.

\begin{figure}[h]
    \centering  
    \includegraphics[width=1\linewidth,page=1]{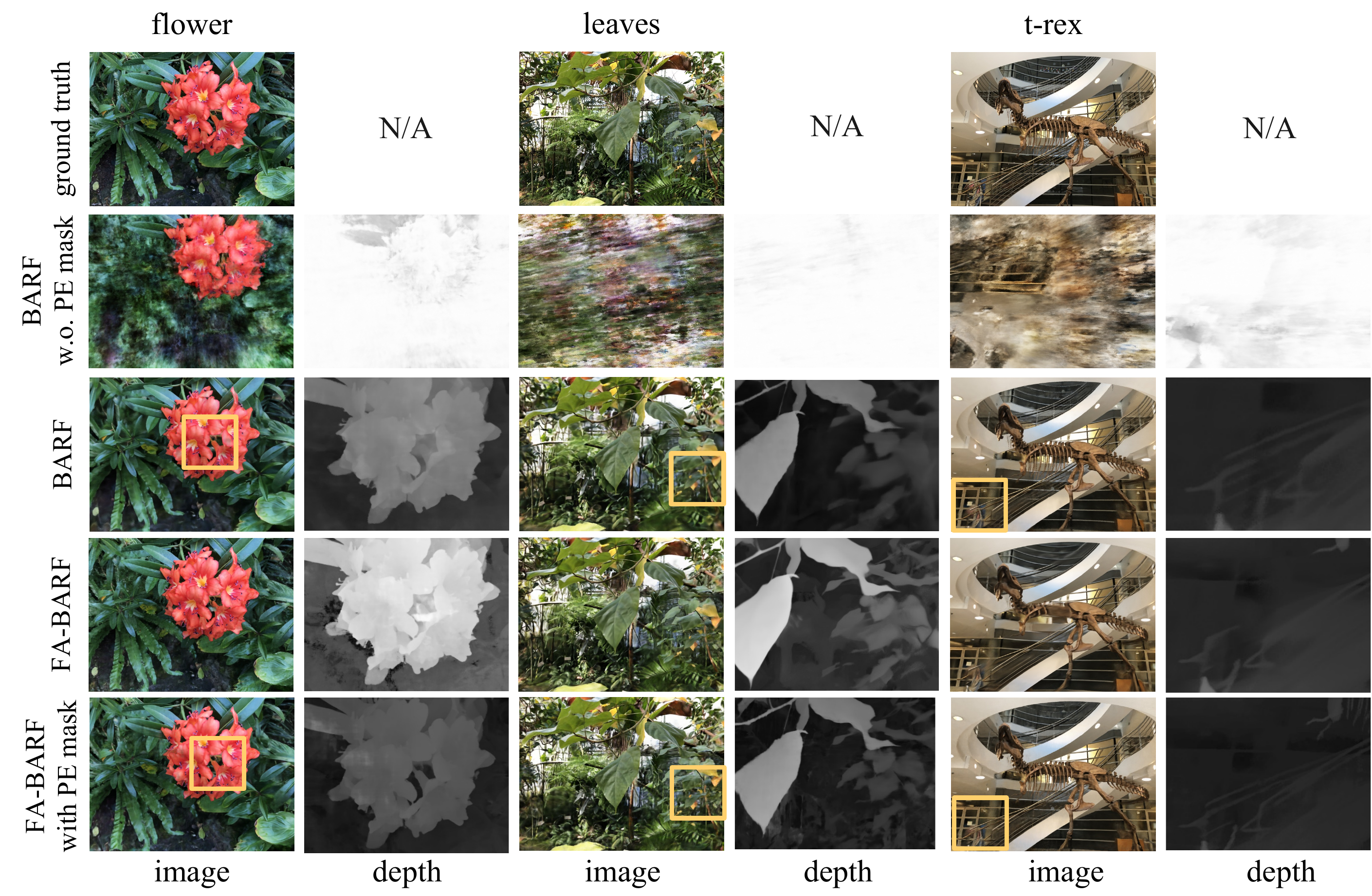}
    \caption{
        Qualitative results of FA-BARF and BARF on real-world scenes from \emph{unknown} camera poses. Compared to original BARF, FA-BARF can capture geometric details marked by the yellow boxes, while BARF has artifacts in depth images.
    }
    \label{fig:nerf-llff}
\end{figure}

\subsubsection{Experimental settings}
The camera poses $\mathbf{p}$ are parameterized with SE(3) following the blender datasets. We initialize all poses with the identity matrix. Considering the complicated nature of real-world scenes compared to object-centric scenes, we compare the performance of FA-BARF and BARF under two settings, with positional encoding mask and without the mask respectively, under the same evaluation criteria described in Sec. \ref{sec:blender-expset}. We find that the camera poses provided in LLFF are also estimated from SfM packages \cite{schonberger2016structure}; therefore, the pose evaluation is at most an indication of how well FA-BARF and BARF agree with classical geometric pose estimation with or without the position encoding annealing strategy.

\subsubsection{Implementation details}
We follow the same architectural settings from the original NeRF \cite{mildenhall2021nerf} and resize the images to $480\times640$ pixels. We train all models for 200K iterations and randomly sample 2048 pixel rays at each optimization step, with a learning rate of $1\times10^{-3}$ for the network $f$ decaying to $1\times10^{-4}$, and $3\times10^{-3}$ for the pose $\mathbf{p}$ decaying to $1\times10^{-5}$. Especially in the setting with positional encoding mask, we linearly open the frequency band gradually for BARF or FA-BARF from iteration 20K to 100K and activate all bands (up to $L = 10$) subsequently.

\subsubsection{Results}
The quantitative results in Table \ref{table:llff-withoutpemask} show that the recovered camera poses from FA-BARF highly agrees with those estimated from off-the-shelf SfM methods, demonstrating the ability of FA-BARF to localize from scratch without the coarse-to-fine process while BARF diverges to incorrect camera poses with poor quality of view synthesis. Furthermore, FA-BARF outperforms in pose registeration and perceptual similarity (LPIPS) with the aid of coarse-to-fine strategy comparing to original BARF as shown in Table \ref{table:llff-withpemask}. This highlights the effectiveness of FA-BARF combining the adapted frequency strategy and coarse-to-fine strategy for joint registration and reconstruction. The qualitative results in Fig. \ref{fig:nerf-llff} show that FA-BARF can capture abundant structure details and geometric information compared to BARF.


\section{Conclusion}\label{conclusion}
In this work, we focused on the task of implicit bundle adjustment, which aims to recover 3D objects or structures as neural radiance models under perturbed or unknown camera poses. We introduced FA-BARF, a frequency-adapted framework for joint optimization of camera poses and 3D NeRF models. Our approach accelerates the training process in object-centric scenes and outperforms BARF without relying on a hand-crafted position encoding mask. We demonstrated that the proposed spatial low-pass filter effectively mitigates the frequency fluctuation phenomenon observed in mainstream papers and optimizes camera poses productively by leveraging uncertainty overlaps.

One limitation of our work is the requirement of a proper frequency band of position encoding to obtain optimal results. Our work can be viewed as a step towards considering implicit bundle adjustment as a fitting problem rather than an overfitting problem, as referred to in the original NeRF. We believe that our work can pave the way for integrating implicit models into real-time applications that demand robust and effective optimization strategies.

In future research, we plan to explore the application of adaptive frequency filters in emerging scene representation technologies, such as 3D Gaussian splatting~\cite{kerbl3Dgaussians} and other related techniques. By extending our approach to these domains, we aim to further enhance the efficiency and effectiveness of 3D reconstruction and rendering pipelines.

\section*{APPENDIX}
\label{APPENDIX}

In this appendix, we illustrate the Jacobians’s derivation of the frequency adapted position encoding $\bar{\gamma}_k$ on $\mathbf{d}_w$, the direction of the a sampled ray in world coordinates and $\mathbf{t}_{c2w}$, the translation of poses in world coordinates in Eq. (\ref{eq:jacobianipe}). 

According to the chain rule, the Jacobian matrix takes the mean $\mu$ and covariance $\Sigma$ of sampled cones  and  as a connection between $\bar{\gamma}_k$ and poses, thus the derivation part related to rotation is
\begin{equation}
    \frac{\partial \bar{\gamma}_k(\mu, \Sigma)}{\partial \mathbf{d}_w} 
    =\frac{\partial  \bar{\gamma}_k(\mu, \Sigma)}{\partial \mu} \cdot  \frac{\partial \mu}{\partial \mathbf{d}_w}+\frac{\partial  \bar{\gamma}_k(\mu, \Sigma)}{\partial \operatorname{diag}\left(\Sigma\right)} \cdot  \frac{\partial \operatorname{diag}(\Sigma)}{\partial \mathbf{d}_w},
\label{eq:ipe_dir}
\end{equation}
and the derivation part related to translation is
\begin{equation}
    \frac{\partial \bar{\gamma}_k(\mu, \Sigma)}{\partial \mathbf{t}_{c2w}} =\frac{\partial \bar{\gamma}_k(\mu, \Sigma)}{\partial \mu}  \cdot \frac{\partial \mu}{\partial \mathbf{t}_{c2w}}.
\label{eq:ipe_trans}
\end{equation}
Futhermore, we unfold the relationship between external parameters composed by rotation $\mathbf{R}_{c2w}$ and translation $\mathbf{t}_{c2w}$ and mean $\mu$ as
\begin{equation}
\boldsymbol{\mu} =\mathbf{t}_{c 2 w}+\boldsymbol{\mu}_t\cdot \mathbf{d}_w,\quad \mathbf{d}_w=\mathbf{R}_{c 2 w}^{T} \cdot \mathbf{d}_c,
\end{equation}
where $\mathbf{d}_w$ satisfies $\|\mathbf{d}_w\|_2^2=1$, and $\mathbf{d}_c$ denotes the ray directions in camera coodinates. Based on this mathematical description, the Jacobian matrix of $\mu$ on $\mathbf{t}_{c2w}$ and $\mathbf{d}_w$ can be calculated as 
\begin{equation}
    \frac{\partial \mu}{\partial \mathbf{d}_w}=\mu_t \cdot \mathbf{I}_3, \quad \frac{\partial \mu}{\partial \mathbf{t}_{c 2 w}}=\mathbf{I}_3.
\label{eq:mean_dir_trans}
\end{equation}
Similarly, the relationship between $\mathbf{d}_{w}$ and covariance $\Sigma$ is 
\begin{equation}
    \operatorname{diag}(\Sigma) =\sigma_t^2\left(\mathbf{d}_w \odot \mathbf{d}_w\right)+\sigma_r^2\left(1-\mathbf{d}_w \odot \mathbf{d}_w\right),
\end{equation}
thus the Jacobian matrix of $ \operatorname{diag}(\Sigma)$ on $\mathbf{d}_w$ can be calculated as 
\begin{equation}
    \begin{aligned}
    \frac{\partial \operatorname{diag}(\Sigma)}{\partial \mathbf{d}_w} & =\left(\sigma_t^2-\sigma_r^2\right) \frac{\partial(\mathbf{d}_w \odot \mathbf{d}_w)}{\partial \mathbf{d}_w} \\
    & =\left(\sigma_t^2-\sigma_r^2\right)\cdot 2\mathbf{d}_w \odot \mathbf{I}_3.
    \end{aligned}
\label{eq:cov_dir}
\end{equation}
According to Eq. (\ref{eq:ipe}), the Jacobian matrix of the frequency adapted position encoding $\bar{\gamma}_k$ related to mean $\mu$ and $\Sigma$ are
\begin{equation}
    \frac{\partial \bar{\gamma}_k(\mu, \Sigma)}{\partial \mu}=2^k\left[\begin{array}{c}\cos \left(\mu\right) \cdot  \exp \left(-\frac{1}{2} \cdot 4^k \cdot\operatorname{diag}\left(\Sigma\right)\right) \odot \mathbf{I}_3\\
    -\sin \left(\mu\right) \cdot  \exp \left(-\frac{1}{2} \cdot 4^k \cdot\operatorname{diag}\left(\Sigma\right)\right)\odot \mathbf{I}_3
    \end{array}\right],
\label{eq:ipe_mean}
\end{equation}
and
\begin{equation}
    \frac{\partial \bar{\gamma}_k(\mu, \Sigma)}{\partial \operatorname{diag}\left(\Sigma\right)}=-2^{2k-1}\left[\begin{array}{l}
    \sin \left(\mu\right) \cdot \exp \left(-\frac{1}{2} \cdot 4^k \cdot\operatorname{diag}\left(\Sigma\right)\right)\odot \mathbf{I}_3 \\
    \cos \left(\mu\right) \cdot \exp \left(-\frac{1}{2} \cdot 4^k \cdot\operatorname{diag}\left(\Sigma\right)\right)\odot \mathbf{I}_3
    \end{array}\right].
\label{eq:ipe_cov}
\end{equation}
Finally, we can obtain the Jacobians’s derivation of $\bar{\gamma}_k$ on $\mathbf{d}_w$ and $\mathbf{t}_{c2w}$ through integrating Eq. (\ref{eq:ipe_mean}), Eq. (\ref{eq:ipe_cov}), Eq. (\ref{eq:mean_dir_trans}) and Eq. (\ref{eq:cov_dir}) into Eq. (\ref{eq:ipe_dir}) and Eq. (\ref{eq:ipe_trans}) respectively. Note that the second part of Eq. (\ref{eq:ipe_dir}) is relatively small compared to the first part in practice, we omit the covariance part in Eq. (\ref{eq:ipe_dir}) to obtain the final expression in Eq. (\ref{eq:jacobianipe}).

\bibliographystyle{IEEEtran}
\bibliography{IEEEabrv, references}
\end{document}